\documentclass[letterpaper]{article} 
\usepackage{aaai24}  
\usepackage{times}  
\usepackage{helvet}  
\usepackage{courier}  
\usepackage[hyphens]{url}  
\usepackage{graphicx} 
\usepackage{natbib}  
\usepackage{caption} 
\usepackage{algorithm}
\usepackage{algorithmic}
\usepackage[table,xcdraw]{xcolor}
\usepackage{newfloat}
\usepackage{listings}
\DeclareCaptionStyle{ruled}{labelfont=normalfont,labelsep=colon,strut=off} 
\floatstyle{ruled}
\newfloat{listing}{tb}{lst}{}
\floatname{listing}{Listing}
\title{TextDiff: Mask-Guided Residual Diffusion Models for Scene Text Image
Super-Resolution}
\author{
    Baolin Liu\equalcontrib\textsuperscript{\rm 1}\textsuperscript{\rm 2},
    Zongyuan Yang\equalcontrib\textsuperscript{\rm 1}\textsuperscript{\rm 2},
    Pengfei Wang\textsuperscript{\rm 1},
    Junjie Zhou\textsuperscript{\rm 1},
    Ziqi Liu\textsuperscript{\rm 1},
    Ziyi Song\textsuperscript{\rm 1},
    Yan Liu\textsuperscript{\rm 1},
    Yongping Xiong\textsuperscript{\rm 1}\thanks{Corresponding Author}
}
\affiliations{
    \textsuperscript{\rm 1}Beijing University of Posts and Telecommunications, Beijing, China, 100876\\
    
    \textsuperscript{\rm 2} \{baolin, yangzongyuan0\}@bupt.edu.cn
}

\usepackage{bibentry}

\begin{document}

\maketitle

\begin{abstract}
The goal of scene text image super-resolution is to reconstruct high-resolution text-line images from unrecognizable low-resolution inputs. The existing methods relying on the optimization of pixel-level loss tend to yield text edges that exhibit a notable degree of blurring, thereby exerting a substantial impact on both the readability and recognizability of the text. To address these issues, we propose TextDiff, the first diffusion-based framework tailored for scene text image super-resolution. It contains two modules: the Text Enhancement Module (TEM) and the Mask-Guided Residual Diffusion Module (MRD). The TEM generates an initial deblurred text image and a mask that encodes the spatial location of the text. The MRD is responsible for effectively sharpening the text edge by modeling the residuals between the ground-truth images and the initial deblurred images. Extensive experiments demonstrate that our TextDiff achieves state-of-the-art (SOTA) performance on public benchmark datasets and can improve the readability of scene text images. Moreover, our proposed MRD module is plug-and-play that effectively sharpens the text edges produced by SOTA methods. This enhancement not only improves the readability and recognizability of the results generated by SOTA methods but also does not require any additional joint training. Available Codes:https://github.com/Lenubolim/TextDiff.
\end{abstract}

\section{Introduction}

Unlike optical character recognition (OCR), scene text image recognition poses persistent challenges owing to factors such as distortion, blurring, and other imaging problems \cite{long2021scene}. Therefore, it is necessary to improve the quality of scene text images.

In the past few years, numerous natural image super-resolution methods \cite{wang2018esrgan,ma2022structure,whang2022deblurring} have been proposed, yet the performance on scene text images remains unsatisfactory. The images produced by these methods exhibit distorted text edges and severe artifacts, which may lead to recognition errors or even render certain text unrecognizable. The key challenge lies in the fact that low-resolution (LR) text images lack crucial textual details, making it arduous to accurately map them to high-resolution (HR) images with significant variations in content, font, and size, solely relying on low-level features. Hence, it becomes crucial to effectively restore text-level information, especially intricate details along the text edges of scene text images.

\begin{figure}[!t]
\centering

\begin{minipage}[r]{0.32\linewidth}
\centering
\includegraphics[width=\linewidth]{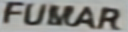}
\captionsetup{labelformat=empty, skip=1pt}
\caption*{\footnotesize TSRN}
\end{minipage}
\hfill
\begin{minipage}[r]{0.32\linewidth}
\centering
\includegraphics[width=\linewidth]{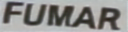}
\captionsetup{labelformat=empty, skip=1pt}
\caption*{\footnotesize TSRN + MRD}
\end{minipage}
\hfill
\begin{minipage}[r]{0.32\linewidth}
\centering
\includegraphics[width=\linewidth]{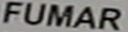}
\captionsetup{labelformat=empty, skip=1pt}
\caption*{\footnotesize HR:FUMAR}
\end{minipage}

\begin{minipage}[r]{0.32\linewidth}
\centering
\includegraphics[width=\linewidth]{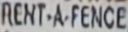}
\captionsetup{labelformat=empty, skip=1pt}
\caption*{\footnotesize TATT}
\end{minipage}
\hfill
\begin{minipage}[r]{0.32\linewidth}
\centering
\includegraphics[width=\linewidth]{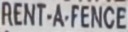}
\captionsetup{labelformat=empty, skip=1pt}
\caption*{\footnotesize TATT + MRD}
\end{minipage}
\hfill
\begin{minipage}[r]{0.32\linewidth}
\centering
\includegraphics[width=\linewidth]{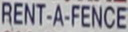}
\captionsetup{labelformat=empty, skip=1pt}
\caption*{\footnotesize HR:RENTAFENCE}
\end{minipage}

\begin{minipage}[r]{0.32\linewidth}
\centering
\includegraphics[width=\linewidth]{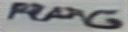}
\captionsetup{labelformat=empty, skip=1pt}
\caption*{\footnotesize C3-STISR}
\end{minipage}
\hfill
\begin{minipage}[r]{0.32\linewidth}
\centering
\includegraphics[width=\linewidth]{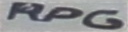}
\captionsetup{labelformat=empty, skip=1pt}
\caption*{\footnotesize C3-STISR + MRD}
\end{minipage}
\hfill
\begin{minipage}[r]{0.32\linewidth}
\centering
\includegraphics[width=\linewidth]{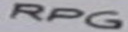}
\captionsetup{labelformat=empty, skip=1pt}
\caption*{\footnotesize HR:RPG}
\end{minipage}

\begin{minipage}[r]{0.32\linewidth}
\centering
\includegraphics[width=\linewidth]{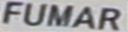}
\captionsetup{labelformat=empty, skip=1pt}
\caption*{\footnotesize TextDiff}
\end{minipage}
\hfill
\begin{minipage}[r]{0.32\linewidth}
\centering
\includegraphics[width=\linewidth]{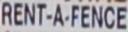}
\captionsetup{labelformat=empty, skip=1pt}
\caption*{\footnotesize TextDiff}
\end{minipage}
\hfill
\begin{minipage}[r]{0.32\linewidth}
\centering
\includegraphics[width=\linewidth]{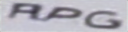}
\captionsetup{labelformat=empty, skip=1pt}
\caption*{\footnotesize TextDiff}
\end{minipage}

\caption{SR results of SOTA methods. The recent text image SR method TSRN, TATT and C3-STISR have achieved progress in terms of quality. However, they still suffer from blurred text and distorted text edges. Our proposed TextDiff can effectively generate sharp and accurate textual images. Our proposed MRD module can effectively sharpen the text edges produced by SOTA methods without any joint training, thereby improving their readability and recognizability.}  
\label{Figure 1}
\end{figure}

To achieve this goal, there are many methods for scene text image super-resolution (STISR) that consider text attributes. The pioneering work, TSRN \cite{wang2020scene} achieves impressive results by capturing sequential character information. Additionally, recent works attempt to incorporate prior knowledge, such as utilizing text priors from recognizers as clues to guide super-resolution. For instance, TPGSR \cite{ma2023text} and TATT \cite{ma2022text} use recognition outputs from CRNN \cite{shi2016end} to guide text reconstruction. TSEPGNet \cite{huang2023text} extracts text embedding and structure priors from upsampled images as auxiliary information to restore clear text.

Despite the considerable progress made by existing methods, there is still room for exploration in STISR. Firstly, text regions deserve greater focus compared to backgrounds. The quality of text region restoration has a substantial impact on text recognition accuracy. Incomplete recovery of certain text segments can lead to recognition errors, especially when dealing with characters that have similar visual features. As shown in Figure\ref{Figure 1}, TATT \cite{ma2022text} fails to entirely restore the content of the text region, such as the letters `R' and `E' in the figure. However, the utilization of text masks encoding location and global structure remains underexplored. Secondly, existing methods suffer from text edge distortion. As discussed in \cite{saharia2022image} and \cite{yang2023docdiff}, ``regression to the mean" can induce text edge distortion in most pixel-loss-based methods. The image restored by TSRN \cite{wang2020scene} can be correctly recognized by recognition models in Figure \ref{Figure 1}. However, from a human visual perception standpoint, the letter 'M' exhibits distortion, and phantom contours are apparent around the letters. Fortunately, diffusion-based methods in natural image super-resolution have shown impressive performance in recovering fine details and reconstructing global structure, without encountering issues like mode collapse or training instability that are commonly seen in Generative Adversarial Networks (GANs) \cite{saharia2022image,li2022srdiff,rombach2022high,ravuri2019classification,gulrajani2017improved}. However, the diffusion model exhibits generative diversity, making it prone to generating images inconsistent with given conditions, thereby posing challenges in generating a fixed text structure \cite{yang2023docdiff}. Additionally, the multi-step sampling in the inference of diffusion models can result in substantial time consumption.

\begin{figure}[t]
\centering
\begin{minipage}[r]{0.32\linewidth}
\centering
\includegraphics[width=\linewidth]{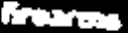}
\end{minipage}
\hfill
\begin{minipage}[r]{0.32\linewidth}
\centering
\includegraphics[width=\linewidth]{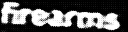}
\end{minipage}
\hfill
\begin{minipage}[r]{0.32\linewidth}
\centering
\includegraphics[width=\linewidth]{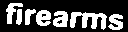}
\end{minipage}

\begin{minipage}[r]{0.32\linewidth}
\centering
\includegraphics[width=\linewidth]{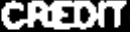}
\end{minipage}
\hfill
\begin{minipage}[r]{0.32\linewidth}
\centering
\includegraphics[width=\linewidth]{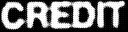}
\end{minipage}
\hfill
\begin{minipage}[r]{0.32\linewidth}
\centering
\includegraphics[width=\linewidth]{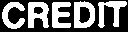}
\end{minipage}

\begin{minipage}[r]{0.32\linewidth}
\centering
\includegraphics[width=\linewidth]{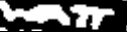}  
\captionsetup{labelformat=empty,skip=2pt}
\caption*{\footnotesize LR mask}
\end{minipage}
\hfill
\begin{minipage}[r]{0.32\linewidth}
\centering
\includegraphics[width=\linewidth]{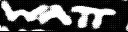}
\captionsetup{labelformat=empty,skip=2pt}
\caption*{\footnotesize Predicted mask}
\end{minipage}
\hfill
\begin{minipage}[r]{0.32\linewidth}
\centering
\includegraphics[width=\linewidth]{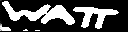}
\captionsetup{labelformat=empty,skip=2pt}
\caption*{\footnotesize HR mask}
\end{minipage}

\caption{Mask images containing global text position information.}  
\label{fig:mask}
\end{figure}

To address these issues above, we propose TextDiff, a novel framework consisting of two modules: the Text Enhancement Module (TEM) and the Mask-Guided Residual Diffusion Module (MRD). The TEM comprises two branches. The first branch integrates semantic information into a coarsely deblurred network, enabling it to capture the global coarse textual structure and yield a coarse output. The second branch focuses on explicit text mask learning. This approach leads to the generation of more accurate masks that exhibit high sensitivity towards textual regions. Examples of the generated text masks are illustrated in Figure \ref{fig:mask}. The MRD module leverages diffusion models to learn the residual distribution between ground-truth images and coarsely deblurred images. Firstly, by predicting the residuals instead of the added noise, we effectively reduce the required number of sampling steps. Secondly, by incorporating text masks and images as input, we provide valuable guidance for residual learning, alleviating text edge distortion and minimizing the generation of diverse outputs. It is worth noting that during inference, we adopt a deterministic sampling scheme to strike a better balance between minimizing distortion and preserving perceptual quality.

Experimental results demonstrate the effectiveness of TextDiff, which shows strong competitiveness with a significant improvement in recognition accuracy compared to the SOTA method, while achieving competitive visual quality on the TextZoom dataset. More importantly, TextDiff still achieves competitive performance with only 5 sampling steps. Ablation experiments demonstrate the effectiveness and necessity of each component in TextDiff. 

Our contributions are summarized as follows:
\begin{itemize}
    \item We introduce TextDiff, which is the first framework in the field of scene text image super-resolution to leverage diffusion models.
    \item TextDiff addresses text edge distortions and blurriness, resulting in more natural image restoration and better preservation of text structure consistency between reconstructed and high-resolution (HR) images. Moreover,  our plug-and-play MRD module effectively enhances the performance of SOTA methods by sharpening the text edges they generate, without additional joint training.
    \item Adequate ablation studies and comparative experiments show that TextDiff achieves SOTA performance on STISR. Additional analysis further confirms the generalizability of TextDiff.
\end{itemize}

\section{Related Works}

\subsection{Single Image Super-Resolution}

Single image super-resolution (SISR) is a fundamental low-level task in computer vision, which goal is to generate HR images from LR inputs. As the seminal work of image super-resolution, SRCNN \cite{dong2014learning} achieves good performance through a simple convolutional network. Besides, in order to further improve the quality of super-resolution (SR) outputs, various methods \cite{liang2021swinir,saharia2022image,chen2023activating} have been proposed. Among them, methods based on the deep generative model, mainly including GAN-based \cite{wang2018esrgan,soh2019natural,wang2021real} and diffusion-based methods \cite{saharia2022image,whang2022deblurring}, have shown convincing image generation ability. However, the method for SISR tasks is not suitable for STISR. The key is that the method of SISR does not consider the structural characteristics of scene text.

\subsection{Scene Text Image Super-Resolution}

Different from SISR, STISR aims to improve the quality of the image while paying attention to the recovery of the text structure. TSRN proposes a real scene text SR dataset, and uses CNN-BiLSTM layers to perceive the sequential information of the text. Following it, \cite{chen2021scene} proposes a Transformer-Based Super-Resolution Network (TBSRN) to extract sequential information, designs a Position-Aware Module and a Content-Aware Module to highlight the position and the content of each character. In addition, TPGSR \cite{ma2023text} obtains the character probability sequence through a text recognition model and merges it with image feature by convolutions. Different from these, by combining the text mask and graphic recognition results of LR  text images, DPMN \cite{zhu2023improving} proposes a plug-and-play module, which can improve the performance of existing models. While these methods have made significant advances, they do not yet fully address issues like text distortion and blurring, which can impair image readlibity. Therefore, our method aims to resolve these specific problems.

\subsection{Diffusion Models}

Diffusion models are latent variable generative frameworks in \cite{sohl2015deep}. Recent work has demonstrated the significant potential of diffusion models in SISR. For example, \cite{li2022srdiff} exploits a Markov chain to convert HR images to latents in simple distribution and then generate SR predictions in the reverse process. However, to the best of our knowledge, diffusion models have not yet been used in STISR. Therefore, in this paper, we explore the performance of diffusion models on STISR for the first time.

\section{Methodology}

In this section, we first provide an overview of the proposed scene text image super-resolution network based on the conditional diffusion model. Then we delve into a detailed description of the working mechanism and role of the conditional diffusion model. We will also introduce the training objective of the proposed network.

\begin{figure*}[!htbp]
\centering
\includegraphics[width=\linewidth, height=!]{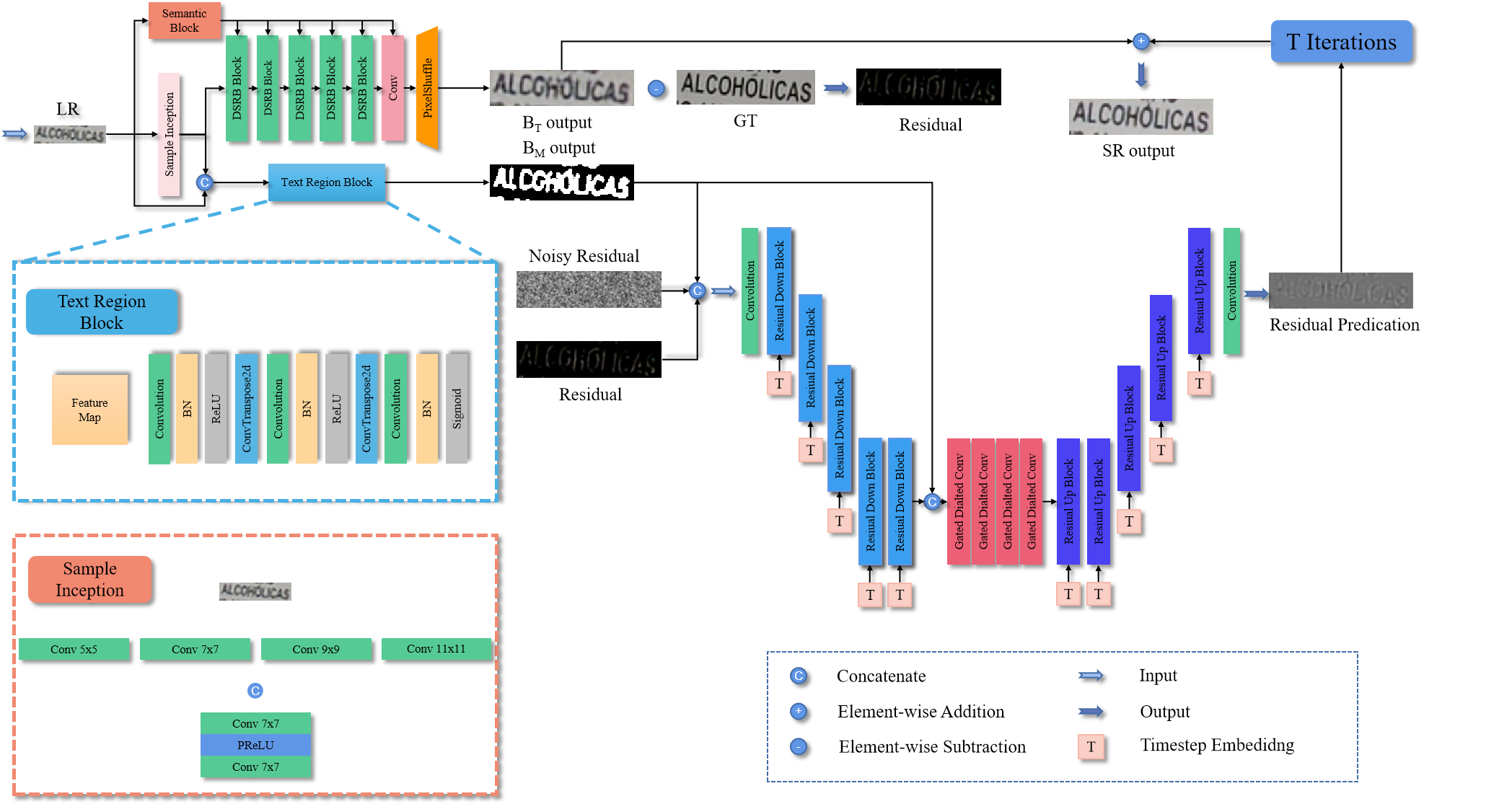}%
\centering
\caption{An overview of the proposed TextDiff for scene text image super-resolution. The TEM module consists of two branches, $\textrm{B}_{T}$ and $\textrm{B}_{M}$, and the U-Net structure is the MRD module.}
\label{img:model}
\centering
\end{figure*}
\subsection{Overall Architecture}

The overall framework of TextDiff is illustrated in Figure \ref{img:model}. First, the LR image is input into the Text Enhancement Module (TEM). The TEM consists of two branches, $\textrm{B}_{T}$ and $\textrm{B}_{M}$. $\textrm{B}_{T}$ outputs a coarsely deblurred image, and $\textrm{B}_{M}$ learns to predict text masks. Then, guided by the mask, the Mask-Guided Residual Diffusion Module (MRD) learns the distribution of residuals between the ground-truth image and the coarsely deblurred image. Finally, this predicted residual is added to the coarsely deblurred image to obtain the final output.

\subsection{Text Enhancement Module}

As shown in Figure \ref{img:model}, the branch $\textrm{B}_{T}$ is a deblurring network incorporated with semantic information. The front part, called Semantic Block, adopts TP Generator (TPG) from TATT \cite{ma2022text} to extract semantic priors, which are then fused with feature maps through TP Interpreter (TPI) in TATT. The addition of this part can effectively integrate the semantic features of the text with the spatial distribution features of the text. The fused results are finally input into the 
Double Sequential Residual Block (DSRB). DSRB introduces wavelet transform on the basis of SRB \cite{wang2020scene} to realize simultaneous learning of spatial and frequency information. This aims to complement the spatial domain details with frequency domain information to generate an output with richer frequency details.

Additionally, the blurred regions in the scene text image mainly focus on the text. We need a text mask to separate the text from the background, and then use the mask to focus the network on the text area. So the branch $\textrm{B}_{M}$ is for mask prediction, implemented by a conventional convolutional network (see Figure \ref{img:model}). And the ground-truth masks are simply generated by calculating the average gray scale of the RGB images.

We define the Gradient Profile Loss as the pixel loss between the $\textrm{B}_{T}$ output and the ground-truth image to recover the approximate information of the global text structure in the LR image:
\begin{equation}\label{Eq.1}
\mathcal{L}_{GP}=\mathrm{E} _{x}||\nabla x_{gt} - \nabla x_{sr}||_{1}
\end{equation}
where $\nabla x_{gt}$ denotes the gradient field of HR images, and $\nabla x_{sr}$ denotes that of SR images. 

Meanwhile, we define the loss for mask learning as dice loss \cite{wang2019shape}. So the $\mathcal{L}_{Mask}$ is used to consider the contour similarity between the $\textrm{B}_{M}$ output $x_{m}$ and the ground-truth text mask $x_{gt_{m}}$. Dice loss can be calculated as the following Eq.\ref{Eq.2} and Eq.\ref{Eq.3}:
\begin{equation}\label{Eq.2}
Dice(P,G)=\frac{2\times  {\textstyle \sum_{x,y}} (P_{x,y}\times G_{x,y})}{\sum_{x,y}(P_{x,y})^{2}+  \sum_{x,y}(G_{x,y})^{2}}
\end{equation}
\begin{equation}\label{Eq.3}
\mathcal{L}_{Mask}=1- Dice(P,G)
\end{equation}
$P_{x,y}$ and $G_{x,y}$ represent the pixel value (x, y) of the predicted
masks and the ground-truths, respectively.

\subsection{Mask-Guided Residual Diffusion Module}

After the LR image is processed by the TEM, $x_{sr}$ and a text mask are obtained. As shown in Figure \ref{img:model}, $x_{sr}$ has restored the general outline and color of the text and other visual features, but compared with the ground-truth image, the edges of the text are still blurred and partially distorted. The residual $x_{res}$ between the $x_{sr}$ and $x_{gt}$ (i.e., $x_{res} = x_{gt} - x_{sr}$) delineates the text outline, as depicted in Figure \ref{img:model}. Thus, we leverage the diffusion model to refine the text contour under the guidance of the text mask to make it more accurate and natural. Experiments show that this residual modeling effectively alleviates the limitations of the regression model in learning text contours, enabling the generation of more perceptually pleasing text. 

Our MRD consists of a diffusion process of progressively adding Gaussian noise and a reverse process of learning residual distributions for denoising.

Specifically, the diffusion process starts from the residual image $x_{res}$, also denoted as $x_0$. 
Then it repeatedly adds Gaussian noise according to the transition kernel $q(x_t \mid x_{t-1})$. And at the maximum time step t, we obtain $x_T$ which is pure Gaussian noise: 
\begin{equation}\label{Eq.6}
q(x_1,...,x_T \mid x_0):=\prod_{t=1}^{T} q(x_t \mid x_{t-1})
\end{equation}
\begin{equation}\label{Eq.7}
q(x_t \mid x_{t-1}):=\mathcal{N} (x_t;\alpha _t x_{t-1},(1-\alpha _t)\mathrm{I})
\end{equation}
The noise schedule $\alpha_{t}$ is a pre-chosen hyperparameter that controls the variance of noise added at each step. Setting $\alpha _t \in  (0, 1)$ for all $t = 1, ...,T$ , $\alpha _{0} = 1 $,  $\bar{\alpha}_{t}= {\textstyle \prod_{i=0}^{t}} \alpha _{i}$, the diffusion process allows sampling $x_t$ at an arbitrary timestep t in closed form:
\begin{equation}\label{Eq.8}
x_{t}(x_0, z) = \sqrt{\bar{\alpha}_{t}} x_t + \sqrt{1 - \bar{\alpha}_{t}} z, z \sim \mathcal{N}(0, \mathrm{I})
\end{equation}
\begin{equation}\label{Eq.9}
q(x_t \mid x_0)=\mathcal{N}(x_t; \sqrt{\bar{\alpha}_t} x_0,(1- \bar{\alpha}_t)\mathrm{I} )
\end{equation}

The reverse process converts noise $x_T$ back into data distribution $x_0$ conditioned on $x_{sr}$ and $x_{m}$. We adopt a deterministic manner to conduct the process:
\begin{equation}\label{Eq.10}
q(x_{t-1} \mid x_{t}, x_0)=\mathcal{N}(x_{t-1}; \mu_t (x_t,x_0),0)
\end{equation}
where $\mu_t (x_t,x_0)$ is calculated as:
\begin{equation}\label{Eq.11}
\mu _t(x_t,x_0)=\sqrt{\bar{\alpha} _{t- 1}} x_0 + \sqrt{1- \bar{\alpha} _{t-1}} \cdot \frac{x_t - \sqrt{\bar{\alpha} _t} x_0}{\sqrt{1- \bar{\alpha} _t} } 
\end{equation}

Furthermore, with $x_0$ as an unknown parameter, the reverse diffusion step can be implemented by substituting the estimate $f_\theta$ in place of $x_0$.
\begin{equation}\label{Eq.12}
p_\theta (x_{t-1} \mid x_t, x_{sr}, x_{m})=q(x_{t-1} \mid x_t, f_\theta (x_t, t, x_{sr}, x_{m}))
\end{equation}

In Eq.\ref{Eq.12}, $x_m$ plays an important role. Looking the U-Net structure in Figure \ref{img:model}, we introduced gated dilated convolution \cite{yang2023gdb}. When processing the input image depth features and text mask, the gating mechanism learns to generate a gating vector that places more attention on the text regions, while dilated convolution enlarges the receptive field. The combination of both allows effective fusion of multi-scale features and enhances the restoration effect for text structure.

Moreover, $f_\theta$ predicts $x_{0}$ instead of noise. The rationale behind this includes: first, cascading both $x_{sr}$ and $x_{m}$ as conditional inputs allows $f_\theta$ to emphasize textual regions when predicting $x_{0}$; second, predicting noise and $x_{0}$ can be converted via Eq.\ref{Eq.8}; third, predicting $x_{0}$ can better exploit $x_{sr}$ and $x_{m}$ as guiding conditions compared to predicting noise, which only depends on $x_{t}$. Meanwhile, predicting noise is more likely to cause diversity that compromises textual structure.

In practice, $f_\theta$ ensures that the learned conditional distribution $p_\theta (x_{t-1} \mid x_t, x_{sr}, x_{m})$ approximates the true reverse diffusion step $q(x_{t-1} \mid x_{t}, x_0)$ as closely as possible. Hence, the training objective is:
\begin{equation}\label{Eq.13}
\mathcal{L}_{\mathrm{DM}}=\mathrm{E}\parallel x_0 - f_{\theta}(\sqrt{\bar{\alpha}_t}x_{res} + \sqrt{1- \bar{\alpha}_t}z,t,x_{sr},x_{m})\parallel _2
\end{equation}
Besides, in order to better restore the contour edges of text, we extract textual structural edge information from the $f_\theta(x_{t},t,x_{sr}, x_{m})$ and $x_{0}$ using a Laplacian kernel $f_{Edge}$ to compute the edge loss. This encourages the model to reconstruct sharper and more coherent textual contours:
\begin{equation}\label{Eq.14}
\mathcal{L}_{Edge} = \mathrm{E} \parallel f_{Edge}(x_{0}) - f_{Edge}(f_\theta(x_{t},t,x_{sr}, x_{m})) \parallel_{2}
\end{equation}

\subsection{Training Objective}

\begin{algorithm}[tb]
\caption{Training}
\label{alg:train}
\textbf{Input}: LR image and its corresponding HR image  pairs $P =\left \{ (x^k_L, x^k_H ) \right \} _{k=1}^K$, the HR mask $x_{{gt}_m}$\\
\textbf{Parameter}: total diffusion step $T$, the predicted mask $x_{m}$, Text Enhancement Module $\textrm{B}_{T}$ and $\textrm{B}_{M}$, denoiser $f_\theta$, noise schedule $\alpha_{0:T}$
\begin{algorithmic}[1] 
\WHILE{not converged}
\STATE Sample $(x_L, x_H) \sim  P$
\STATE $x_{sr}$ = $\textrm{B}_{T}(x_L)$, compute $x_{res}$ = $x_{H} -  x_{L}$
\STATE $x_{m}$ = $\textrm{B}_{M}(x_L)$
\STATE Sample $z \sim \mathcal{N} (0, \mathrm{I})$, and $t \sim \mathrm{Uniform}(\{1, . . ., T \})$
\STATE Take a gradient step on\\
\hspace{4mm} $x_t = \sqrt{\bar{\alpha} _t}x_{res}+ \sqrt{1-\bar{\alpha} _t}z$,\\
\hspace{4mm} $\mathcal{L}_{total}(\textrm{B}_{T},\textrm{B}_{M},f_{\theta};x_H,x_{sr},x_{m},x_{gt_m},x_{res})$
\ENDWHILE
\end{algorithmic}
\end{algorithm}

\begin{algorithm}[tb]
\caption{Inference}
\label{alg:infer}
\textbf{Input}: LR image $x_L$, total diffusion step $T$\\
\textbf{Parameter}: Text Enhancement Module $\textrm{B}_{T}$ and $\textrm{B}_{M}$, denoiser $f_\theta$, noise schedule $\alpha_{0:T}$, the predicted mask $x_{m}$\\
\textbf{Output}: High-resolution image generated by TextDiff
\begin{algorithmic}[1] 
\STATE Sample $(x_T) \sim \mathcal{N} (0, \mathrm{I})$
\STATE $x_{sr}$ = $\textrm{B}_{T}(x_L)$
\STATE $x_{m}$ = $\textrm{B}_{M}(x_L)$
\STATE \textbf{for} $t = T : 1$ \textbf{do}
\STATE \hspace{4mm}$x_{res} = f_\theta(x_{sr}, x_{m}, x_t, t)$,
\STATE \hspace{4mm}$x_{t-1}=\sqrt{\bar{\alpha} _{t-1}}x_{res}+\frac{\sqrt{1-\bar{\alpha} _{t-1}}(x_t-\sqrt{\bar{\alpha} _{t}}x_{res})}{\sqrt{1-\bar{\alpha} _{t}}}$
\STATE \textbf{end for}
\STATE \textbf{return} $x_{sr} + x_0$ as SR prediction
\end{algorithmic}
\end{algorithm}

The overall loss function for training the network is comprised of three parts: the loss from the TEM, the loss from the MRD, and a joint loss $\mathcal{L}_{joint}$ combining both modules. 
\begin{equation}\label{Eq.15}
\mathcal{L}_{TEM} = \lambda_1 \mathcal{L}_{GP} + \lambda_2 \mathcal{L}_{Mask} + \mathcal{L}_{TP}
\end{equation}
\begin{equation}\label{Eq.16}
\mathcal{L}_{MRD} = \mathcal{L}_{Edge} + \mathcal{L}_{\mathrm{DM}}
\end{equation}
where $\mathcal{L}_{TP}$ is a text prior loss for TPG \cite{ma2022text}, $\lambda_1$ and $\lambda_2$ are hyperparameters.
Specifically, the joint loss of the two modules is formulated as:
\begin{equation}\label{Eq.17}
\mathcal{L}_{joint}= \sum_{i=1}^{N} \mathrm{E} \parallel R_i(x_{sr}  + f_\theta(x_{t},t,x_{sr}, x_{m})) - R_i(x_{gt})\parallel_1
\end{equation}
where $R_i$ denotes the feature map output from the CRNN \cite{shi2016end} after the i-th activation layer. Since CRNN is an optical character recognition model capable of perceiving textual patterns, we utilize its publicly available pre-trained weights to extract features without additional training on TextZoom. Incorporating this CRNN-based loss allows the network to better restore textual content.

Thus, the overall loss function is
\begin{equation}\label{Eq.18}
\mathcal{L}_{total} = \mathcal{L}_{TEM} + \mathcal{L}_{MRD} + \lambda \mathcal{L}_{joint}
\end{equation}
where $\lambda$ is a hyperparameter.

The training and inference procedures are presented in Algorithm \ref{alg:train} and Algorithm \ref{alg:infer}, respectively.

\section{Experiments and Results}

\subsection{Datasets and Implementation Details}

We conduct training and evaluation on the TextZoom \cite{wang2020scene} which is collected in real-world scenarios. This dataset consists of 17,367 LR-HR image pairs in the training set. The test set is divided into easy, medium, and hard subsets comprising 1,619, 1,411, and 1,343 LR-HR pairs respectively, based on the camera focal length. The size of LR images is $16 \times 64$, while the size of HR images is $32 \times 128$.

Our model is implemented with PyTorch 2.0 deep learning library and all the experiments are conducted on one RTX 4090 GPU. AdamW \cite{loshchilov2017decoupled} is utilized as the optimizer with a learning rate of $1 \times 10^{-4}$, and the batch size is set to 16. The total time steps $T$ are set to 200. The number of training iterations is one million. We use a linear increase in $\beta_{1:T}$ from $1 \times 10^{-6}$ to $1 \times  10^{-2}$, $\alpha _t = 1 - \beta_t$. Moreover, the two stages of the proposed network are trained jointly, where the weight coefficient $\lambda$ for the joint loss is set to 5. The weights for $\mathcal{L}_{GP}$ and $\mathcal{L}_{Mask}$ are set to 0.5 and 3 in Eq.\ref{Eq.15}, respectively. Additional model configuration information is given in the supplementary material.

\subsection{Metrics and Experimental Results}

\begin{table*}[!htbp]
\resizebox{\linewidth}{!}{%
\begin{tabular}{c|cccc|cccc|cccc}
\hline
& \multicolumn{4}{c|}{ASTER\cite{shi2018aster}} & \multicolumn{4}{c|}{MORAN\cite{luo2019moran}} & \multicolumn{4}{c}{CRNN\cite{shi2016end}} \\
\hline  
Method & Easy & Medium & Hard & Average & Easy & Medium & Hard & Average & Easy & Medium & Hard & Average \\
\hline 
BICUBIC & 67.4\% & 42.4\% & 31.2\% & 48.2\% & 60.6\% & 37.9\% & 30.8\% & 44.1\% & 36.4\% & 21.1\% & 21.1\% & 26.8\%\\

SRCNN,TPAMI2015 & 70.6\% & 44.0\% & 31.5\% & 50.0\% & 63.9\% & 40.0\% & 29.4\% & 45.6\% &  41.1\% & 22.3\% & 22.0\% & 29.2\%\\
 
SRResNet,CVPR2017 & 69.4\% & 50.5\% & 35.7\% & 53.0\% & 66.0\% & 47.1\% & 33.4\% & 49.9\% &  45.2\% & 32.6\% & 25.5\% & 35.1\%\\

TSRN,ECCV2020 & 73.4\% & 56.3\% & 40.1\% & 57.7\% & 70.1\% & 53.3\% & 37.9\% & 54.8\% & 54.8\% & 41.3\% & 32.3\% & 43.6\%\\
 
TBSRN,CVPR2021 & 75.7\% & 59.9\% & 41.6\% & 60.0\% & 74.1\% & 57.0\% & 40.8\% & 58.4\% & 59.6\% & 47.1\% & 35.3\% & 48.1\% \\
 
PCAN,MM2021 & 77.5\% & 60.7\% & 43.1\% & 61.5\% & 73.7\% & 57.6\% & 41.0\% & 58.5\% & 59.6\% & 45.4\% & 34.8\% & 47.4\% \\
 
TPGSR,TIP2023 & 77.0\% & 60.9\% & 42.4\% & 60.9\% & 72.2\% & 57.8\% & 41.3\% & 57.8\% & 61.0\% & 49.9\% & 36.7\% & 49.8\%\\
 
TPGSR-3,TIP2023 & 78.9\% & 62.7\% & 44.5\% & 62.8\% & 74.9\% & 60.5\% & 44.1\% & 60.5\% &  63.1\% & 52.0\% & 38.6\% & 51.8\%\\

DocDiff,MM2023 & 69.8\% & 56.2\% & 39.8\% & 56.2\% & 64.5\% & 50.5\% & 35.6\% & 51.2\% &  52.5\% & 41.3\% & 31.2\% & 42.4\%\\

TATT,CVPR2022 & 78.5\% & 63.3\% & 45.0\% & 63.3\% & 72.9\% &  61.0\% & 43.8\% & 60.1\% &  64.3\% & 54.2\% & 39.1\% & 53.3\%\\
 
C3-STISR,IJCAI2022 & 79.1\% & 63.3\% & 46.8\% & 64.1\% & 74.2\% &  61.0\% & 43.2\% &  60.5\% & 65.2\% &  53.6\% & 39.8\% & 53.7\%\\

DPMN(+TATT),AAAI2023 & 79.3\% & 64.1\% & 45.2\% &  63.9\% & 73.3\% &  61.5\% & 43.9\% &  60.5\% & 64.4\% & 54.2\% & 39.3\% & 53.4\%\\

TSAN,AAAI2023 & 79.6\% & 64.1\% &  45.3\% &  64.1\% & 78.4\% &  61.3\% &  45.1\% &  62.7\% &  64.6\% &  53.3\% &  38.8\% &  53.0\%\\

STNet,TMM2024 & 80.8\% & 63.8\% &  45.8\% &  64.6\% & 77.7\% &  61.9\% &  44.2\% &  62.3\% &  62.2\% &  51.3\% &  37.5\% &  51.1\%\\

\textbf{Ours(TextDiff-5)} & $79.8\%$ &  $65.0\%$ &  $47.2\%$ &$65.0\%$ & $76.9\%$ & $61.7\%$ & $44.4\%$ &  $62.0\%$ & 64.7\% & $54.9\%$ &$40.2\%$ &$54.0\%$\\

\textbf{Ours(TextDiff-200)} & $81.3\% $ & $66.5\%$ & $48.7\%$ & $66.6\%$ & $77.7\%$ & $63.1\%$ & $45.0\%$ & $63.0\%$ & $64.8\%$ & $55.9\%$ & $40.2\%$ & $54.3\%$ \\
\hline
HR & 94.2\% & 87.7\% & 76.2\% & 86.6\%  & 91.2\% & 85.3\% & 74.2\% & 84.1\% & 76.4\% & 75.1\% & 64.6\% & 72.4\%\\
\hline

\end{tabular}%
}
\caption{Comparison with state-of-the-art SR methods on three subsets of the TextZoom testsets. `-3' means multi-stage settings in \cite{ma2023text}. TextDiff-n means applying n-step sampling($T$).}
\label{tab:baseline}
\end{table*}

\begin{figure*}[tb]
\centering

\begin{minipage}[r]{0.06\linewidth}
\centering
\textbf {\footnotesize Bicubic}
\end{minipage}
\hfill
\begin{minipage}[r]{0.15\linewidth}
\centering
\includegraphics[width=\linewidth]{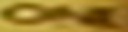}%
\end{minipage}
\hfill
\begin{minipage}[r]{0.15\linewidth}
\centering
\includegraphics[width=\linewidth]{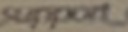}%
\end{minipage}
\hfill
\begin{minipage}[r]{0.15\linewidth}
\centering
\includegraphics[width=\linewidth]{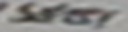}%
\end{minipage}
\hfill
\begin{minipage}[r]{0.15\linewidth}
\centering
\includegraphics[width=\linewidth]{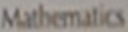}%
\end{minipage}
\hfill
\begin{minipage}[r]{0.15\linewidth}
\centering
\includegraphics[width=\linewidth]{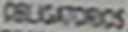}%
\end{minipage}
\hfill
\begin{minipage}[r]{0.15\linewidth}
\centering
\includegraphics[width=\linewidth]{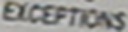}%
\end{minipage}

\begin{minipage}[r]{0.06\linewidth}
\centering
\textbf{\footnotesize TSRN}
\end{minipage}
\hfill
\begin{minipage}[r]{0.15\linewidth}
\centering
\includegraphics[width=\linewidth]{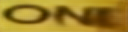}%
\end{minipage}
\hfill
\begin{minipage}[r]{0.15\linewidth}
\centering
\includegraphics[width=\linewidth]{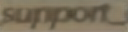}%
\end{minipage}
\hfill
\begin{minipage}[r]{0.15\linewidth}
\centering
\includegraphics[width=\linewidth]{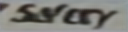}%
\end{minipage}
\hfill
\begin{minipage}[r]{0.15\linewidth}
\centering
\includegraphics[width=\linewidth]{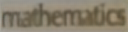}%
\end{minipage}
\hfill
\begin{minipage}[r]{0.15\linewidth}
\centering
\includegraphics[width=\linewidth]{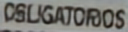}%
\end{minipage}
\hfill
\begin{minipage}[r]{0.15\linewidth}
\centering
\includegraphics[width=\linewidth]{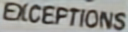}%
\end{minipage}

\begin{minipage}[r]{0.06\linewidth}
\centering
\textbf {\footnotesize TATT}
\end{minipage}
\hfill
\begin{minipage}[r]{0.15\linewidth}
\centering
\includegraphics[width=\linewidth]{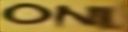}%
\end{minipage}
\hfill
\begin{minipage}[r]{0.15\linewidth}
\centering
\includegraphics[width=\linewidth]{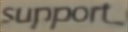}%
\end{minipage}
\hfill
\begin{minipage}[r]{0.15\linewidth}
\centering
\includegraphics[width=\linewidth]{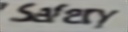}%
\end{minipage}
\hfill
\begin{minipage}[r]{0.15\linewidth}
\centering
\includegraphics[width=\linewidth]{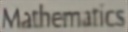}%
\end{minipage}
\hfill
\begin{minipage}[r]{0.15\linewidth}
\centering
\includegraphics[width=\linewidth]{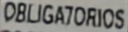}%
\end{minipage}
\hfill
\begin{minipage}[r]{0.15\linewidth}
\centering
\includegraphics[width=\linewidth]{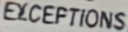}%
\end{minipage}

\begin{minipage}[r]{0.06\linewidth}
\centering
\textbf{\footnotesize C3-STISR}
\end{minipage}
\hfill
\begin{minipage}[r]{0.15\linewidth}
\centering
\includegraphics[width=\linewidth]{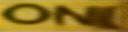}%
\end{minipage}
\hfill
\begin{minipage}[r]{0.15\linewidth}
\centering
\includegraphics[width=\linewidth]{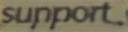}%
\end{minipage}
\hfill
\begin{minipage}[r]{0.15\linewidth}
\centering
\includegraphics[width=\linewidth]{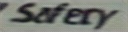}%
\end{minipage}
\hfill
\begin{minipage}[r]{0.15\linewidth}
\centering
\includegraphics[width=\linewidth]{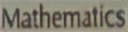}%
\end{minipage}
\hfill
\begin{minipage}[r]{0.15\linewidth}
\centering
\includegraphics[width=\linewidth]{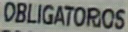}%
\end{minipage}
\hfill
\begin{minipage}[r]{0.15\linewidth}
\centering
\includegraphics[width=\linewidth]{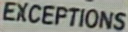}%
\end{minipage}

\begin{minipage}[r]{0.06\linewidth}
\centering
\textbf{\footnotesize Ours}
\end{minipage}
\hfill
\begin{minipage}[r]{0.15\linewidth}
\centering
\includegraphics[width=\linewidth]{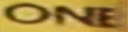}%
\end{minipage}
\hfill
\begin{minipage}[r]{0.15\linewidth}
\centering
\includegraphics[width=\linewidth]{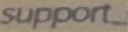}%
\end{minipage}
\hfill
\begin{minipage}[r]{0.15\linewidth}
\centering
\includegraphics[width=\linewidth]{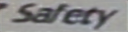}%
\end{minipage}
\hfill
\begin{minipage}[r]{0.15\linewidth}
\centering
\includegraphics[width=\linewidth]{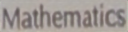}%
\end{minipage}
\hfill
\begin{minipage}[r]{0.15\linewidth}
\centering
\includegraphics[width=\linewidth]{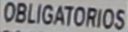}%
\end{minipage}
\hfill
\begin{minipage}[r]{0.15\linewidth}
\centering
\includegraphics[width=\linewidth]{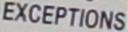}%
\end{minipage}

\begin{minipage}[r]{0.06\linewidth}
\centering
\textbf{\footnotesize HR}
\end{minipage}
\hfill
\begin{minipage}[r]{0.15\linewidth}
\centering
\includegraphics[width=\linewidth]{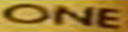}%
\end{minipage}
\hfill
\begin{minipage}[r]{0.15\linewidth}
\centering
\includegraphics[width=\linewidth]{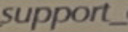}%
\end{minipage}
\hfill
\begin{minipage}[r]{0.15\linewidth}
\centering
\includegraphics[width=\linewidth]{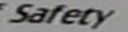}%
\end{minipage}
\hfill
\begin{minipage}[r]{0.15\linewidth}
\centering
\includegraphics[width=\linewidth]{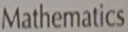}%
\end{minipage}
\hfill
\begin{minipage}[r]{0.15\linewidth}
\centering
\includegraphics[width=\linewidth]{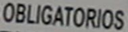}%
\end{minipage}
\hfill
\begin{minipage}[r]{0.15\linewidth}
\centering
\includegraphics[width=\linewidth]{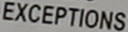}%
\end{minipage}

\caption{The SR image results on TextZoom.}  
\label{Figure:3}
\end{figure*}

\begin{table}[tb]
\centering
\begin{tabular}{c|cccc}
\hline
  &  \multicolumn{4}{c}{Metric}  \\
\hline
Method & PSRN$\uparrow$  & SSIM$\uparrow$  & LPIPS$\downarrow$ & MANIQA$\uparrow$ \\
\hline
BICUBIC & 20.35 & 0.6961 & 0.2199 & 0.4413\\

TSRN &  21.42 & 0.7691 & 0.0973& 0.4533\\

TATT & 21.52 & 0.7930 & 0.0945& 0.4493\\

C3-STISR & \textbf{21.60} & \textbf{0.7931} & 0.0944 & 0.4521\\

\hline
TextDiff & 21.52 & 0.7841& \textbf{0.0822} &  \textbf{0.4589}\\
\hline
\end{tabular}
\caption{Evaluation of competitive STISR models on three subsets of the TextZoom testset. The bold numbers denote the best score.}
\label{tab:PSNR}
\end{table}

Following the common practice in TSRN, to measure the downstream task performance, we calculate the text recognition accuracy for the text recognition task. Additionally, We evaluate the quality of the SR images using metrics including the no-reference image quality assessment method MANIQA \cite{yang2022maniqa}, the full-reference method LPIPS \cite{zhang2018unreasonable}, PSNR and SSIM.

We evaluate our TextDiff and compare it with existing super-resolution models, including SRCNN \cite{dong2015image}, SRResNet \cite{ledig2017photo}, TSRN \cite{wang2020scene}, TBSRN \cite{chen2021scene}, PCAN \cite{zhao2021scene}, TPGSR \cite{ma2023text}, DocDiff \cite{yang2023docdiff}, TATT \cite{ma2022text}, C3-STISR \cite{zhao2022c3}, DPMN \cite{zhu2023improving}, TSAN~\cite{zhu2023gradient} and STNet~\cite{zhao2024scene}. 

As shown in Table \ref{tab:baseline}, we can clearly observe that our TextDiff achieves a significant improvement in accuracy compared to existing methods. On average, our method improves recognition accuracy by 2.0\%, 0.3\% and 0.6\% on ASTER, MORAN and CRNN, respectively. Furthermore, TextDiff outperforms all existing methods in recognition accuracy with only 5-step sampling (see the quantitative results taking other sampling steps in the supplementary material.). The consistent accuracy gains on multiple models validate the effectiveness of our TextDiff. For comprehensiveness, we also compare TextDiff with SOTA diffusion-based document enhancement method (DocDiff), and the results show the superiority of our method for STISR. 

Additionally, as shown in Figure \ref{Figure:3}, most methods have insufficient capability in recovering textual structures, whereas our method can effectively alleviate this problem. We also give quantitative results for image quality evaluation, as shown in Table \ref{tab:PSNR}. Our TextDiff achieves the best MANIQA and LPIPS metrics, while also achieving competitive PSNR and SSIM. Notably, we obtain the LPIPS of 0.0822, a 12\% reduction compared to C3-STISR and a 10\% reduction compared to TATT.

Finally, we list the failure cases of TextDiff and our future work in the supplementary material.

\subsection{Ablation Studies}

\begin{table}[tb]
\centering
\begin{tabular}{c|cccc}
\hline
Method &  Easy & Medium & Hard & Average \\
\hline
TextDiff w/o $\textrm{B}_{M}$ & 77.1 & 60.2 & 44.3 & 61.8\\

TextDiff with NU & 79.5 & 64.0 & 46.9 & 64.5\\

TextDiff with NP & 76.6 & 61.1 & 43.3 & 61.4\\

TextDiff w/o $\mathcal{L}_{joint}$ & 79.8 & 63.6 & 47.2 & 64.6\\
\hline
TextDiff & 81.3 & 66.5 & 48.7 & 66.6\\
\hline
\end{tabular}
\caption{Quantitative ablation study results on TextZoom. And the recognition accuracy (\%) of different methods are based on ASTER. ``w/o" denotes without, ``NP" denotes ``Noise Prediction", ``NU" denotes an equivalent U-Net is used in place of a diffusion model.}
\label{tab:ablation}
\end{table}

In this section, we perform ablation studies to analyze the contribution of different motivations and model components. All the evaluations are validated on TextZoom. The quantitative and qualitative results of the ablation experiments are provided in Table \ref{tab:ablation} and in the supplementary material, respectively.

\subsubsection{The Mask Branch.} 
To validate the efficacy of the proposed text mask branch, we ablate it by removing the text mask branch and comparing performance with and without it. Results in Table \ref{tab:ablation} show that the text mask branch does play a role in improving image quality restoration. The text mask branch enables the network to focus more on the features within the text regions, not only facilitating better extraction of text-level characteristics but also enhancing the connectivity between the two modules. Besides, since the output of the text mask branch does not directly optimize the scene text image, the parameter increase incurred by this branch is not the reason for improved super-resolution performance.

\subsubsection{The MRD.} 
To validate that the performance improvement of our proposed MRD module is not solely due to an increase in parameter size, we replace the MRD module with an identical U-Net structure to form a two-stage regression model. Experimental results show that while simply cascading more encoder-decoder layers can improve recognition accuracy, the perceptual quality is still poor and some fonts are distorted. Cascading the MRD module can improve the mean recognition accuracy by 2.1$\%$. This verifies the validity of MRD module.

\subsubsection{Noise Prediction and Residual Learning.} 
We employ the MRD module to predict noise and perform original stochastic sampling. As shown in Table \ref{tab:ablation}, for STISR, predicting residuals from more known conditions achieves better performance than predicting noise. It is shown that prediction residuals are beneficial to text recovery along with deterministic sampling.

\subsubsection{Perceptual Loss.}  
From Table \ref{tab:ablation}, we can see that the perceptual loss can improve accuracy by over 1.9\%. The reason is that the perceptual loss can accurately focus on the difference between text and background, and calculate the loss from shallow textual structure features to deep textual semantic features in the image. In addition, the predicted value input to the loss function is $x_{sr} + x_0$, which can implicitly adjust the fusion effect of pixel-wise addition of TEM and MRD. 

\subsection{Extensions}

\begin{table}[tb]
\centering
\begin{tabular}{c|cccc}
\hline
Method &  Easy & Medium & Hard & Average \\
\hline
TSRN & 73.4 & 56.3 & 40.1 & 57.7\\
TSRN + MRD & \textbf{74.3} & \textbf{59.3} & \textbf{42.4} & \textbf{59.7}\\
\hline
TATT & 78.5 & 63.3 & 45.0 & 63.3\\
TATT + MRD & \textbf{79.0} & \textbf{63.5} & \textbf{46.3} & \textbf{64.0}\\
\hline
C3-STISR & 79.1 & 63.3 & 46.8 & 64.1\\
C3-STISR + MRD & \textbf{79.3} & \textbf{64.2} & 46.8 & \textbf{64.4}\\
\hline
HR & 94.2 & 87.7 & 76.2 & 86.6\\
\hline
\end{tabular}
\caption{Quantitative results of cascading MRD module with other methods on TextZoom. And the recognition accuracy (\%) of different methods are based on ASTER. The bold numbers denote the better score between the baseline and improved method by MRD.}
\label{tab:extension}
\end{table}

We explore cascading our MRD module with other mainstream approaches to further validate its efficacy and extensibility. Specifically, we achieve joint inference without any joint training, simply by direct inference. The inference results are shown in Table \ref{tab:extension} and Figure \ref{Figure 1} (see additional cascade inference results in the supplementary material.). The results sufficiently demonstrate that the effectiveness of MRD comes not only from the algorithm itself, but also from its ability to complement other types of methods. For instance, from the cascaded results of MRD with TSRN, it is evident that MRD effectively complements the incomplete font structure obtained from TSRN, leading to results that are more aligned with human perception. As a result, the recognition accuracy of TSRN increases by 2\%, without any additional training. Overall, embedding MRD as a post-processing module into existing pipelines can provide further performance boost. 

\subsection{Discussions}

\begin{figure}[tb]
\centering

\begin{minipage}[r]{0.12\linewidth}
\centering
\textbf {\footnotesize Bicubic}\\
\hfill
\end{minipage}
\hfill
\begin{minipage}[r]{0.42\linewidth}
\centering
\includegraphics[width=\linewidth]{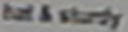}
\captionsetup{labelformat=empty,skip=2pt}
\caption*{\footnotesize{\textcolor{red}{wiitra}y /P:18.94 /S:0.46}}
\end{minipage}
\hfill
\begin{minipage}[r]{0.42\linewidth}
\centering
\includegraphics[width=\linewidth]{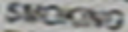}
\captionsetup{labelformat=empty,skip=2pt}
\caption*{\footnotesize s\textcolor{red}{w}o\textcolor{red}{od} /P:16.24 /S:0.45}
\end{minipage}

\begin{minipage}[r]{0.12\linewidth}
\centering
\textbf{\footnotesize TSRN}\\
\hfill
\end{minipage}
\hfill
\begin{minipage}[r]{0.42\linewidth}
\centering
\includegraphics[width=\linewidth]{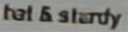}  
\captionsetup{labelformat=empty,skip=2pt}
\caption*{\footnotesize ha\textcolor{red}{lf}st\textcolor{red}{a}rdy /P:19.30 /S:0.65}
\end{minipage}
\hfill
\begin{minipage}[r]{0.42\linewidth}
\centering
\includegraphics[width=\linewidth]{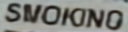}
\captionsetup{labelformat=empty,skip=2pt}
\caption*{\footnotesize s\textcolor{red}{iv}o\textcolor{red}{io}n\textcolor{red}{o} /P:20.50 /S:0.78}
\end{minipage}

\begin{minipage}[r]{0.12\linewidth}
\centering
\textbf{\footnotesize TATT}\\
\hfill
\end{minipage}
\hfill
\begin{minipage}[r]{0.42\linewidth}
\centering
\includegraphics[width=\linewidth]{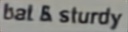}
\captionsetup{labelformat=empty,skip=2pt}
\caption*{\footnotesize hat\&sturdy /P:20.44 /S:0.69}
\end{minipage}
\hfill
\begin{minipage}[r]{0.42\linewidth} 
\centering
\includegraphics[width=\linewidth]{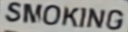}
\captionsetup{labelformat=empty,skip=2pt}
\caption*{\footnotesize smoking /P:18.49 /S:0.80}
\end{minipage}

\begin{minipage}[r]{0.12\linewidth}
\centering
\textbf{\footnotesize C3-STISR}\\
\hfill
\end{minipage}
\hfill
\begin{minipage}[r]{0.42\linewidth}
\centering
\includegraphics[width=\linewidth]{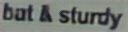} 
\captionsetup{labelformat=empty,skip=2pt}
\caption*{\footnotesize hat\textcolor{red}{d}sturdy /P:18.03 /S:0.64}
\end{minipage}
\hfill
\begin{minipage}[r]{0.42\linewidth}
\centering
\includegraphics[width=\linewidth]{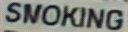}
\captionsetup{labelformat=empty,skip=2pt}
\caption*{\footnotesize s\textcolor{red}{n}oking /P:20.60 /S:0.79}
\end{minipage}

\begin{minipage}[r]{0.12\linewidth}
\centering
\textbf{\footnotesize Ours}\\
\hfill
\end{minipage}
\hfill
\begin{minipage}[r]{0.42\linewidth}
\centering
\includegraphics[width=\linewidth]{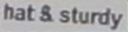}
\captionsetup{labelformat=empty,skip=2pt}
\caption*{\footnotesize hat\&sturdy /P:16.96 /S:0.74}
\end{minipage}
\hfill
\begin{minipage}[r]{0.42\linewidth}
\centering 
\includegraphics[width=\linewidth]{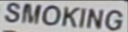}
\captionsetup{labelformat=empty,skip=2pt}
\caption*{\footnotesize smoking /P:16.49 /S:0.82}
\end{minipage}

\begin{minipage}[r]{0.12\linewidth}
\centering
\textbf{\footnotesize HR}\\
\hfill
\end{minipage}
\hfill
\begin{minipage}[r]{0.42\linewidth}
\centering
\includegraphics[width=\linewidth]{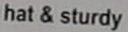}
\captionsetup{labelformat=empty,skip=2pt}
\caption*{\footnotesize hat\&sturdy}
\end{minipage}
\hfill
\begin{minipage}[r]{0.42\linewidth}
\centering 
\includegraphics[width=\linewidth]{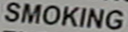}
\captionsetup{labelformat=empty,skip=2pt}
\caption*{\footnotesize smoking}
\end{minipage}

\caption{PSNR, SSIM and Recognition results of estimated HR images and real HR images. ``P" and ``S" represent PSNR and SSIM.}  
\label{Figure:4}
\end{figure}

We mainly discuss the applicability of PSNR and SSIM metrics for the STISR. As shown in Figure \ref{Figure:4}, we find that for phrases such as ``hat\&sturdy", the PSNR and SSIM of images obtained by using existing methods are relatively high, but the recovered fonts are distorted or blurred. In contrast, the image recovered by TextDiff can perfectly restore the font structure, but the PSNR and SSIM are low. Whether it is from the perspective of human perception or model recognition results, our method is relatively good. This also confirms the quantitative results in Table \ref{tab:PSNR}. In summary, as discussed in \cite{wang2020scene} and \cite{yang2023docdiff}, it can be concluded that PSNR and SSIM metrics only partially align with human perception when applied to text images.

\section{Conclusions}

In this paper, we propose TextDiff, a novel framework for scene text image super-resolution. The framework comprises two key components: the Text Enhancement Module (TEM) and the Mask-Guided Residual Diffusion Module (MRD). The TEM includes two branches, one branch realizes the preliminary deblurring of images combined with semantic information, and the other branch realizes the learning of text masks. The MRD module learns the residual distribution under the guidance of the text mask, and further restores the text structure and edge contour. Extensive experiments demonstrate that compared to existing methods, our approach achieves state-of-the-art (SOTA) performance on public benchmark datasets. In addition, our MRD module is plug-and-play that effectively sharpens the text edges produced by SOTA methods. We believe our work will provide valuable intuition for further improvement of the STISR task. 

\bibliography{Formatting-Instructions-LaTeX-2024}

\appendix

\section{Model Configuration}

We set the initial number of Channels in U-net to 64, with channel multipliers 1, 2, 4, 4. The Residual Down Block and Residual Up Block are structurally consistent. They are composed of convolutional layers, activation layers, dropout layers, self-attention layer, and normalization layers. The only difference is the number of channels. The dropout is set to 0.1.

\section{More Details In Ablation Study}

\begin{figure}[t]
\centering

\begin{minipage}[r]{0.12\linewidth}
\centering
\textbf {LR}\\
\vfill
\end{minipage}
\hfill
\begin{minipage}[r]{0.4\linewidth}
\centering
\includegraphics[width=\linewidth]{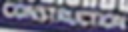}
\end{minipage}
\hfill
\begin{minipage}[r]{0.4\linewidth}
\centering
\includegraphics[width=\linewidth]{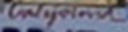}
\end{minipage}
\\[1ex]

\begin{minipage}[r]{0.12\linewidth}
\centering
\textbf{TextDiff with NU}\\
\vfill
\end{minipage}
\hfill
\begin{minipage}[r]{0.4\linewidth}
\centering
\includegraphics[width=\linewidth]{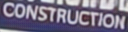}  
\end{minipage}
\hfill
\begin{minipage}[r]{0.4\linewidth}
\centering
\includegraphics[width=\linewidth]{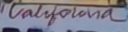}
\end{minipage}
\\[1ex]

\begin{minipage}[r]{0.12\linewidth}
\centering
\textbf{TextDiff with NP}\\
\vfill
\end{minipage}
\hfill
\begin{minipage}[r]{0.4\linewidth}
\centering
\includegraphics[width=\linewidth]{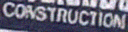}  
\end{minipage}
\hfill
\begin{minipage}[r]{0.4\linewidth}
\centering
\includegraphics[width=\linewidth]{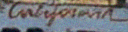}
\end{minipage}
\\[1ex]

\begin{minipage}[r]{0.12\linewidth}
\centering
\textbf{TextDiff w/o $\mathrm{B}_M$}\\
\vfill
\end{minipage}
\hfill
\begin{minipage}[r]{0.4\linewidth}
\centering
\includegraphics[width=\linewidth]{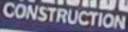}
\end{minipage}
\hfill
\begin{minipage}[r]{0.4\linewidth} 
\centering
\includegraphics[width=\linewidth]{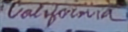}
\end{minipage}
\\[1ex]

\begin{minipage}[r]{0.12\linewidth}
\centering
\textbf{TextDiff w/o $L_{joint}$}\\
\vfill
\end{minipage}
\hfill
\begin{minipage}[r]{0.4\linewidth}
\centering
\includegraphics[width=\linewidth]{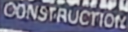}
\end{minipage}
\hfill
\begin{minipage}[r]{0.4\linewidth} 
\centering
\includegraphics[width=\linewidth]{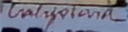}
\end{minipage}
\\[1ex]

\begin{minipage}[r]{0.12\linewidth}
\centering
\textbf{TextDiff}\\
\vfill
\end{minipage}
\hfill
\begin{minipage}[r]{0.4\linewidth}
\centering
\includegraphics[width=\linewidth]{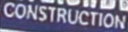}
\end{minipage}
\hfill
\begin{minipage}[r]{0.4\linewidth}
\centering 
\includegraphics[width=\linewidth]{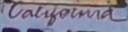}
\end{minipage}
\\[1ex]

\begin{minipage}[r]{0.12\linewidth}
\centering
\textbf{HR}\\
\vfill
\end{minipage}
\hfill
\begin{minipage}[r]{0.4\linewidth}
\centering
\includegraphics[width=\linewidth]{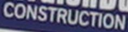}
\end{minipage}
\hfill
\begin{minipage}[r]{0.4\linewidth}
\centering 
\includegraphics[width=\linewidth]{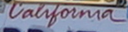}
\end{minipage}

\caption{Qualitative ablation study results on TextZoom. “w/o” denotes without, “NP” denotes
“Noise Prediction”, “NU” denotes an equivalent U-Net is used in place of a diffusion model.}  
\label{Figure:ABLATION}
\end{figure}

In Figure \ref{Figure:ABLATION}, we present the qualitative results of several ablation experiments. We can draw the following analysis conclusions:

\begin{itemize}
    \item Adding encoder-decoder layers to optimize pixel loss in a cascade manner may not necessarily improve legibility.
    \item Predicting the added noise while sampling too few steps may result in a sampled image with noise and blurred text edges.
    \item If the text mask information is not used, the edges of the text in the generated image will be more blurred, which reflects that the text mask has a positive effect on text restoration.
    \item Perceptual loss plays an important role in the recovery of text structure.
\end{itemize}

\section{Discussion About Sampling Steps}

\begin{table}[!htbp]
\centering
\begin{tabular}{c|cccc}
\hline
Method &  Easy & Medium & Hard & Average \\
\hline

TextDiff-5 & 79.8 & 65.0 & 47.2 & 65.0\\

TextDiff-20 & 79.9 & 65.3 & 47.3 & 65.2\\

TextDiff-50  & 80.0 & 65.3 & 47.5 & 65.3\\

TextDiff-110  & 80.2 & 65.5 & 47.7 & 65.5\\

TextDiff-200 & 81.3 & 66.5 & 48.7 & 66.6\\
\hline
\end{tabular}
\caption{Quantitative results for different sampling steps on TextZoom. TextDiff-n means applying n-step sampling ($T$).}
\label{tab:sample}
\end{table}

We set the number of sampling steps to 200 in the training process. In order to further explore the influence of different sampling steps on STISR, we conduct experiments with different sampling steps, and the number of sampling steps is set to 5, 20, 50, 110 and 200 respectively. The quantitative results of different sampling steps are shown in Table \ref{tab:sample}. As expected, as we increase the sampling step, the recognition accuracy is gradually increasing. We emphasize that TextDiff is able to produce high-quality images within a few steps, thanks to its training strategy of predicting original data and its deterministic sampling strategy.

\section{More Details In Cascading Inference}

\begin{figure*}[!t]
\centering

\begin{minipage}[r]{0.1\linewidth}
\centering
\textbf {TSRN}\\
\hfill
\end{minipage}
\hfill
\begin{minipage}[r]{0.2\linewidth}
\centering
\includegraphics[width=\linewidth]{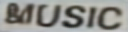}
\captionsetup{labelformat=empty,skip=2pt}
\caption*{}
\end{minipage}
\hfill
\begin{minipage}[r]{0.2\linewidth}
\centering
\includegraphics[width=\linewidth]{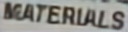}
\captionsetup{labelformat=empty,skip=2pt}
\caption*{}
\end{minipage}
\hfill
\begin{minipage}[r]{0.2\linewidth}
\centering
\includegraphics[width=\linewidth]{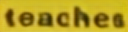}
\captionsetup{labelformat=empty,skip=2pt}
\caption*{}
\end{minipage}
\hfill
\begin{minipage}[r]{0.2\linewidth}
\centering
\includegraphics[width=\linewidth]{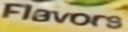}
\captionsetup{labelformat=empty,skip=2pt}
\caption*{}
\end{minipage}

\begin{minipage}[r]{0.1\linewidth}
\centering
\textbf {TSRN + MRD}\\
\hfill
\end{minipage}
\hfill
\begin{minipage}[r]{0.2\linewidth}
\centering
\includegraphics[width=\linewidth]{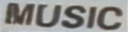}
\captionsetup{labelformat=empty,skip=2pt}
\caption*{}
\end{minipage}
\hfill
\begin{minipage}[r]{0.2\linewidth}
\centering
\includegraphics[width=\linewidth]{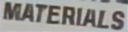}
\captionsetup{labelformat=empty,skip=2pt}
\caption*{}
\end{minipage}
\hfill
\begin{minipage}[r]{0.2\linewidth}
\centering
\includegraphics[width=\linewidth]{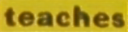}
\captionsetup{labelformat=empty,skip=2pt}
\caption*{}
\end{minipage}
\hfill
\begin{minipage}[r]{0.2\linewidth}
\centering
\includegraphics[width=\linewidth]{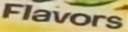}
\captionsetup{labelformat=empty,skip=2pt}
\caption*{}
\end{minipage}

\begin{minipage}[r]{0.1\linewidth}
\centering
\textbf {TATT}\\
\hfill
\end{minipage}
\hfill
\begin{minipage}[r]{0.2\linewidth}
\centering
\includegraphics[width=\linewidth]{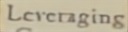}
\captionsetup{labelformat=empty,skip=2pt}
\caption*{}
\end{minipage}
\hfill
\begin{minipage}[r]{0.2\linewidth}
\centering
\includegraphics[width=\linewidth]{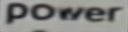}
\captionsetup{labelformat=empty,skip=2pt}
\caption*{}
\end{minipage}
\hfill
\begin{minipage}[r]{0.2\linewidth}
\centering
\includegraphics[width=\linewidth]{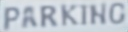}
\captionsetup{labelformat=empty,skip=2pt}
\caption*{}
\end{minipage}
\hfill
\begin{minipage}[r]{0.2\linewidth}
\centering
\includegraphics[width=\linewidth]{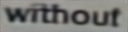}
\captionsetup{labelformat=empty,skip=2pt}
\caption*{}
\end{minipage}

\begin{minipage}[r]{0.1\linewidth}
\centering
\textbf {TATT \ + MRD}\\
\hfill
\end{minipage}
\hfill
\begin{minipage}[r]{0.2\linewidth}
\centering
\includegraphics[width=\linewidth]{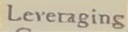}
\captionsetup{labelformat=empty,skip=2pt}
\caption*{}
\end{minipage}
\hfill
\begin{minipage}[r]{0.2\linewidth}
\centering
\includegraphics[width=\linewidth]{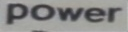}
\captionsetup{labelformat=empty,skip=2pt}
\caption*{}
\end{minipage}
\hfill
\begin{minipage}[r]{0.2\linewidth}
\centering
\includegraphics[width=\linewidth]{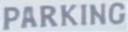}
\captionsetup{labelformat=empty,skip=2pt}
\caption*{}
\end{minipage}
\hfill
\begin{minipage}[r]{0.2\linewidth}
\centering
\includegraphics[width=\linewidth]{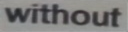}
\captionsetup{labelformat=empty,skip=2pt}
\caption*{}
\end{minipage}

\begin{minipage}[r]{0.1\linewidth}
\centering
\textbf {C3-STISR}\\
\hfill
\end{minipage}
\hfill
\begin{minipage}[r]{0.2\linewidth}
\centering
\includegraphics[width=\linewidth]{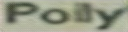}
\captionsetup{labelformat=empty,skip=2pt}
\caption*{}
\end{minipage}
\hfill
\begin{minipage}[r]{0.2\linewidth}
\centering
\includegraphics[width=\linewidth]{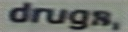}
\captionsetup{labelformat=empty,skip=2pt}
\caption*{}
\end{minipage}
\hfill
\begin{minipage}[r]{0.2\linewidth}
\centering
\includegraphics[width=\linewidth]{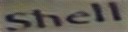}
\captionsetup{labelformat=empty,skip=2pt}
\caption*{}
\end{minipage}
\hfill
\begin{minipage}[r]{0.2\linewidth}
\centering
\includegraphics[width=\linewidth]{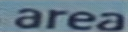}
\captionsetup{labelformat=empty,skip=2pt}
\caption*{}
\end{minipage}

\begin{minipage}[r]{0.1\linewidth}
\centering
\textbf {C3-STISR + MRD}\\
\hfill
\end{minipage}
\hfill
\begin{minipage}[r]{0.2\linewidth}
\centering
\includegraphics[width=\linewidth]{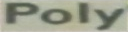}
\captionsetup{labelformat=empty,skip=2pt}
\caption*{}
\end{minipage}
\hfill
\begin{minipage}[r]{0.2\linewidth}
\centering
\includegraphics[width=\linewidth]{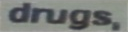}
\captionsetup{labelformat=empty,skip=2pt}
\caption*{}
\end{minipage}
\hfill
\begin{minipage}[r]{0.2\linewidth}
\centering
\includegraphics[width=\linewidth]{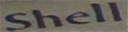}
\captionsetup{labelformat=empty,skip=2pt}
\caption*{}
\end{minipage}
\hfill
\begin{minipage}[r]{0.2\linewidth}
\centering
\includegraphics[width=\linewidth]{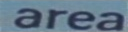}
\captionsetup{labelformat=empty,skip=2pt}
\caption*{}
\end{minipage}

\caption{Some qualitative results of cascade inferences.}  
\label{Figure:joint}
\end{figure*}

The specific implementation of the cascade operation is to first input the LR image into the existing method for processing, and then input the obtained image into the Mask-Guided Residual Diffusion Module (MRD) we proposed. In particular, since joint training is not required, in the experiments, for existing methods, we directly use models from the official implementation or perform the identical hyper-parameters as reported in the official implementations to train the baseline models. Furthermore, we present some additional qualitative results of cascade inferences in Figure \ref{Figure:joint}. From these figures, we can intuitively feel the strong recovery ability and robustness of our proposed MRD module.

\section{Failure Case and Limitation}

\begin{figure*}[!t]
\centering

\begin{minipage}[r]{0.06\linewidth}
\centering
\textbf {LR}\\
\vfill
\end{minipage}
\hfill
\begin{minipage}[r]{0.3\linewidth}
\centering
\includegraphics[width=\linewidth]{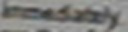}
\captionsetup{labelformat=empty,skip=2pt}
\caption*{}
\end{minipage}
\hfill
\begin{minipage}[r]{0.3\linewidth}
\centering
\includegraphics[width=\linewidth]{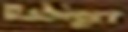}
\captionsetup{labelformat=empty,skip=2pt}
\caption*{}
\end{minipage}
\hfill
\begin{minipage}[r]{0.3\linewidth}
\centering
\includegraphics[width=\linewidth]{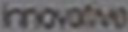}
\captionsetup{labelformat=empty,skip=2pt}
\caption*{}
\end{minipage}

\begin{minipage}[r]{0.06\linewidth}
\centering
\textbf {C3-STISR}\\
\vfill
\end{minipage}
\hfill
\begin{minipage}[r]{0.3\linewidth}
\centering
\includegraphics[width=\linewidth]{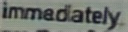}
\captionsetup{labelformat=empty,skip=2pt}
\caption*{\footnotesize immed\textcolor{red}{i}ately}
\end{minipage}
\hfill
\begin{minipage}[r]{0.3\linewidth}
\centering
\includegraphics[width=\linewidth]{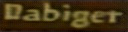}
\captionsetup{labelformat=empty,skip=2pt}
\caption*{\footnotesize \textcolor{red}{D}abige\textcolor{red}{t}}
\end{minipage}
\hfill
\begin{minipage}[r]{0.3\linewidth}
\centering
\includegraphics[width=\linewidth]{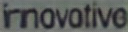}
\captionsetup{labelformat=empty,skip=2pt}
\caption*{\footnotesize \textcolor{red}{r}nov\textcolor{red}{o}tive}
\end{minipage}

\begin{minipage}[r]{0.06\linewidth}
\centering
\textbf {TextDiff}\\
\vfill
\end{minipage}
\hfill
\begin{minipage}[r]{0.3\linewidth}
\centering
\includegraphics[width=\linewidth]{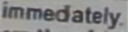}
\captionsetup{labelformat=empty,skip=2pt}
\caption*{immed\textcolor{red}{i}ately}
\end{minipage}
\hfill
\begin{minipage}[r]{0.3\linewidth}
\centering
\includegraphics[width=\linewidth]{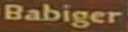}
\captionsetup{labelformat=empty,skip=2pt}
\caption*{\footnotesize \textcolor{red}{B}biger}
\end{minipage}
\hfill
\begin{minipage}[r]{0.3\linewidth}
\centering
\includegraphics[width=\linewidth]{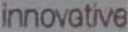}
\captionsetup{labelformat=empty,skip=2pt}
\caption*{\footnotesize innov\textcolor{red}{o}tive}
\end{minipage}

\begin{minipage}[r]{0.06\linewidth}
\centering
\textbf {HR}\\
\vfill
\end{minipage}
\hfill
\begin{minipage}[r]{0.3\linewidth}
\centering
\includegraphics[width=\linewidth]{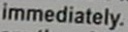}
\captionsetup{labelformat=empty,skip=2pt}
\caption*{\footnotesize immediately}
\end{minipage}
\hfill
\begin{minipage}[r]{0.3\linewidth}
\centering
\includegraphics[width=\linewidth]{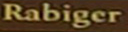}
\captionsetup{labelformat=empty,skip=2pt}
\caption*{\footnotesize Rabiger}
\end{minipage}
\hfill
\begin{minipage}[r]{0.3\linewidth}
\centering
\includegraphics[width=\linewidth]{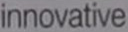}
\captionsetup{labelformat=empty,skip=2pt}
\caption*{\footnotesize innovative}
\end{minipage}

\caption{Some failure cases of our proposed model.}  
\label{Figure:error}
\end{figure*}

Figure \ref{Figure:error} shows some failure cases. For some LR images, due to the similarity between characters, there is a problem that some letters are restored to other characters. This problem appears to be less likely to occur compared to existing methods. However, although the corresponding strategy (e.g., residual learning) is used in our proposed TextDiff to solve this problem, it is not enough. Therefore, we leave it as future work.

\section{Future Work}

In our work, there are still problems to be solved. As in the experiment, it is found that character substitution still exists in scene text recovery, which is also an inherent problem in existing methods. In TextDiff, the potential solutions are to enhance the text positioning ability, optimize the conditional input method of the diffusion model, increase the diversity of samples, etc.

\end{document}